\documentclass[sigconf, screen]{acmart}
\usepackage{multirow}
\usepackage{graphicx}
\usepackage{pifont}
\usepackage[table,dvipsnames,svgnames]{xcolor} 
\definecolor{sea}{RGB}{226,234,243}

\AtBeginDocument{%
  }

\setcopyright{acmlicensed}
\copyrightyear{2026}
\acmYear{2026}
\acmDOI{XXXXXXX.XXXXXXX}

\acmISBN{XXX-X-XXXX-XXXX-X/2026/XX}

\begin{document}

\title{MASRA: MLLM-Assisted Semantic-Relational Consistent Alignment for Video Temporal Grounding}

\author{Ran Ran}
\affiliation{%
  \institution{School of Computer Science and Engineering, University of Electronic Science and Technology of China}
  \city{Chengdu}
  \country{China}
}

\author{Jiwei Wei}
\affiliation{%
  \institution{School of Computer Science and Engineering, University of Electronic Science and Technology of China}
  \city{Chengdu}
  \country{China}
}

\author{Shuchang Zhou}
\affiliation{%
  \institution{School of Computer Science and Engineering, University of Electronic Science and Technology of China}
  \city{Chengdu}
  \country{China}
}

\author{Yitong Qin}
\affiliation{%
  \institution{School of Computer Science and Engineering, University of Electronic Science and Technology of China}
  \city{Chengdu}
  \country{China}
}

\author{Shiyuan He}
\affiliation{%
  \institution{School of Computer Science and Engineering, University of Electronic Science and Technology of China}
  \city{Chengdu}
  \country{China}
}

\author{Zeyu Ma}
\affiliation{%
  \institution{School of Computer Science and Engineering, University of Electronic Science and Technology of China}
  \city{Chengdu}
  \country{China}
}

\author{Yuyang Zhou}
\affiliation{%
  \institution{School of Cyberspace Security, Hainan University}
  \city{Haikou}
  \country{China}
}

\author{Yang Yang}
\affiliation{%
  \institution{School of Computer Science and Engineering, University of Electronic Science and Technology of China}
  \city{Chengdu}
  \country{China}
}

\renewcommand{\shortauthors}{Anonymous.}

\begin{abstract}
Video Temporal Grounding (VTG) faces a cross-modal semantic gap that often leads to background features being incorrectly aligned with the query, while directly matching the query to moments results in insufficient discriminability and consistency of temporal semantics. 
To address this issue, we propose \textbf{M}LLM-\textbf{A}ssisted \textbf{S}emantic-\textbf{R}elational Consistent \textbf{A}lignment (MASRA), a training-time MLLM-based optimization framework for VTG. 
MASRA leverages an MLLM during training to produce two forms of textual priors, namely event-level descriptions with temporal spans and clip-level captions, and instantiates two MLLM-assisted alignments. Event Semantic Temporal Alignment (ESTA) aligns temporal context with event semantics to explicitly strengthen the correspondence between semantics and temporal events and improve span-level separability.
Local Relational Consistency Alignment (LRCA) constructs a textual relation matrix derived from clip-level captions and aligns it with the temporal feature similarity matrix in the model, enhancing temporal consistency while capturing local structural information.
MASRA includes two simple supporting modules, semantic-guided enhancement and second-order relational attention, to better utilize the learned semantic context and relational structure.
Moreover, we introduce Decoupled Alignment Interaction (DAI) with a context-aware codebook to adaptively absorb query-irrelevant semantics and alleviate the cross-modal gap.
The MLLM is only invoked during training and is not used at inference. Extensive experiments show that MASRA outperforms existing methods, and ablation studies validate its effectiveness.
\end{abstract}

%%
%% The code below is generated by the tool at http://dl.acm.org/ccs.cfm.
%% Please copy and paste the code instead of the example below.
%%


%%
%% Keywords. The author(s) should pick words that accurately describe
%% the work being presented. Separate the keywords with commas.
%% A "teaser" image appears between the author and affiliation
%% information and the body of the document, and typically spans the
%% page.

% \received{20 February 2007}
% \received[revised]{12 March 2009}
% \received[accepted]{5 June 2009}

%%
%% This command processes the author and affiliation and title
%% information and builds the first part of the formatted document.
\maketitle

\section{Introduction}
\label{sec:intro}
\begin{figure}[t]
	\begin{center}
		{\includegraphics[width=0.99\linewidth]{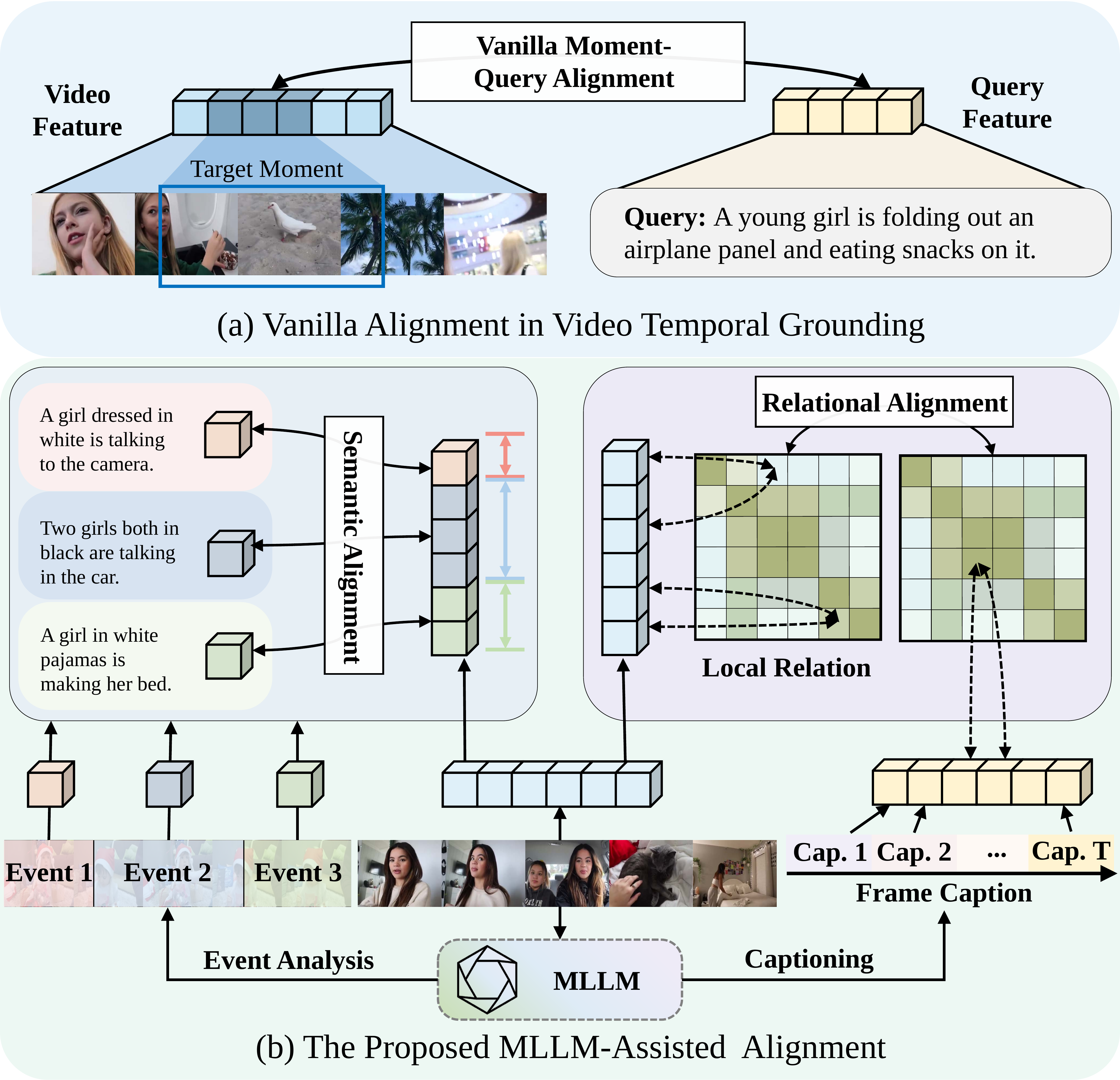}}
		\caption{(a) Vanilla VTG alignment, where the query is directly matched to raw video features. (b) The proposed MASRA 
        leverages MLLM-generated textual priors to drive event-level semantic alignment and local relational alignment in training, thereby bridging the cross-modal gap and strengthening boundary cues.}
		\label{head}
	\end{center} 
\end{figure}
Video has become the dominant medium on the internet, and users increasingly expect to retrieve specific moments with natural-language queries. This trend makes Video Temporal Grounding (VTG) an important topic in multimodal understanding ~\cite{anne2017localizing,liu2018cross,yuan2019semantic,zhang2020learning}. The task aims to localize, in an untrimmed video, the temporal segment that best matches a given query, which serves both retrieval and understanding, and also acts as a fundamental capability for higher-level applications~\cite{zhang2023temporal,moon2023query}.

Early methods can be roughly divided into two main lines~\cite{zhang2021multi,li2022compositional}: one aligns sentences with predefined temporal proposals for selection~\cite{chen2020rethinking,lu2019debug,wang2022negative,liu2022skimming}; the other directly predicts temporal spans through cross-modal interactions between language and clips~\cite{anne2017localizing,gao2017tall,zhang2020span}. Subsequently, detection transformer frameworks have been introduced into VTG, using query-based proposal detection to effectively eliminate hand-crafted proposal components~\cite{lei2021detecting,jang2023knowing,sun2024tr}. These families of approaches have achieved significant improvements in overall grounding performance.

However, existing methods still face notable challenges in cross-modal alignment and boundary representation~\cite{zhang2023temporal}. 
First, the modality gap remains pronounced. Because of the discrepancy between textual queries and visual semantics, direct alignment tends to pull query-irrelevant background segments toward the text space, thereby reducing local discriminability~\cite{lee2024bam,wang2025diffusion}. 
Moreover, many approaches primarily focus on the correlation between the query and the target moment, as illustrated in Figure~\ref{head}(a), which leads to insufficient semantic separability from the surrounding temporal context. 
In particular, temporal-segment discriminability and local relational consistency remain relatively weak, making it difficult to form clear and stable boundary cues~\cite{wu2024clim,wuclipself,gu2025breaking}.

Meanwhile, the prevailing training paradigm lacks dense semantic supervision and local relational supervision. Fortunately, current multimodal large language models (MLLM) demonstrate reliable understanding and description at the image level and short event level~\cite{liu2023visual,wang2024qwen2}, and can accurately generate two types of textual priors: clip-level detailed captions and event analyses. We leverage these priors as additional supervision: aligning language semantics with temporal events and aligning local relations at the clip level.  Textual information thus acts as an alignment mediator to bridge the modality gap while simultaneously strengthening boundary discrimination and temporal consistency, as shown in Figure~\ref{head}(b).

On this basis, we propose \textbf{M}LLM-\textbf{A}ssisted \textbf{S}emantic-\textbf{R}elational Consistent \textbf{A}lignment (MASRA), a novel training-time supervision framework for video temporal grounding. The core idea is to convert sparse grounding supervision into two denser forms of MLLM-generated textual guidance, including event-level semantic priors and clip-level relational priors. Accordingly, MASRA introduces two complementary objectives, Event Semantic Temporal Alignment (ESTA) and Local Relational Consistency Alignment (LRCA). ESTA aligns temporally pooled context with event descriptions to strengthen semantic-temporal correspondence and span-level discriminability. LRCA regularizes the clip-level visual-similarity structure using a textual relation graph derived from clip-level captions, thereby improving local temporal consistency and boundary separability. These alignments are used only during training and incur no inference-time MLLM cost.

In the model backbone, we design Decoupled Alignment Interaction (DAI), which uses a context-aware codebook to allow the model to adaptively absorb query-irrelevant semantics during cross-modal interaction, thereby stabilizing the alignment process. In addition, to better integrate the proposed supervision, we employ two simple yet effective modules. Semantic-Guided Enhancement (SGE) uses semantically aligned temporal context to guide feature fusion, and Second-Order Relational Attention (SORA) performs lightweight local refinement on the learned similarity map to sharpen temporal structure. Together, SGE and SORA serve as supporting components that help the backbone better absorb the proposed event-level and relation-level priors.

Our main contributions are summarized as follows:

\begin{itemize}
    \item We introduce two complementary alignments supervised by MLLM-generated textual priors. ESTA aligns temporal context with event semantics to strengthen semantic-temporal correspondence and improve span-level discriminability, while LRCA aligns local relational structures by fitting a textual relation graph to the visual similarity matrix, thereby enhancing temporal consistency and boundary separability. Both alignments are used only during training and introduce no inference-time overhead.
    
    \item  We design a MASRA framework for VTG that incorporates Decoupled Alignment Interaction based on a context-aware codebook for adaptive cross-modal alignment. We further include two supporting modules, semantic-guided enhancement for context-guided feature fusion and second-order relational attention for local refinement of the similarity map, to help the backbone better exploit the proposed event-level and relation-level priors.

    \item Extensive experiments demonstrate the effectiveness of our method. On multiple VTG benchmarks, MASRA achieves consistent improvements and competitive performance over strong existing methods.
\end{itemize}

\begin{figure*}[t]
  \centering
   \includegraphics[width=0.99\linewidth]{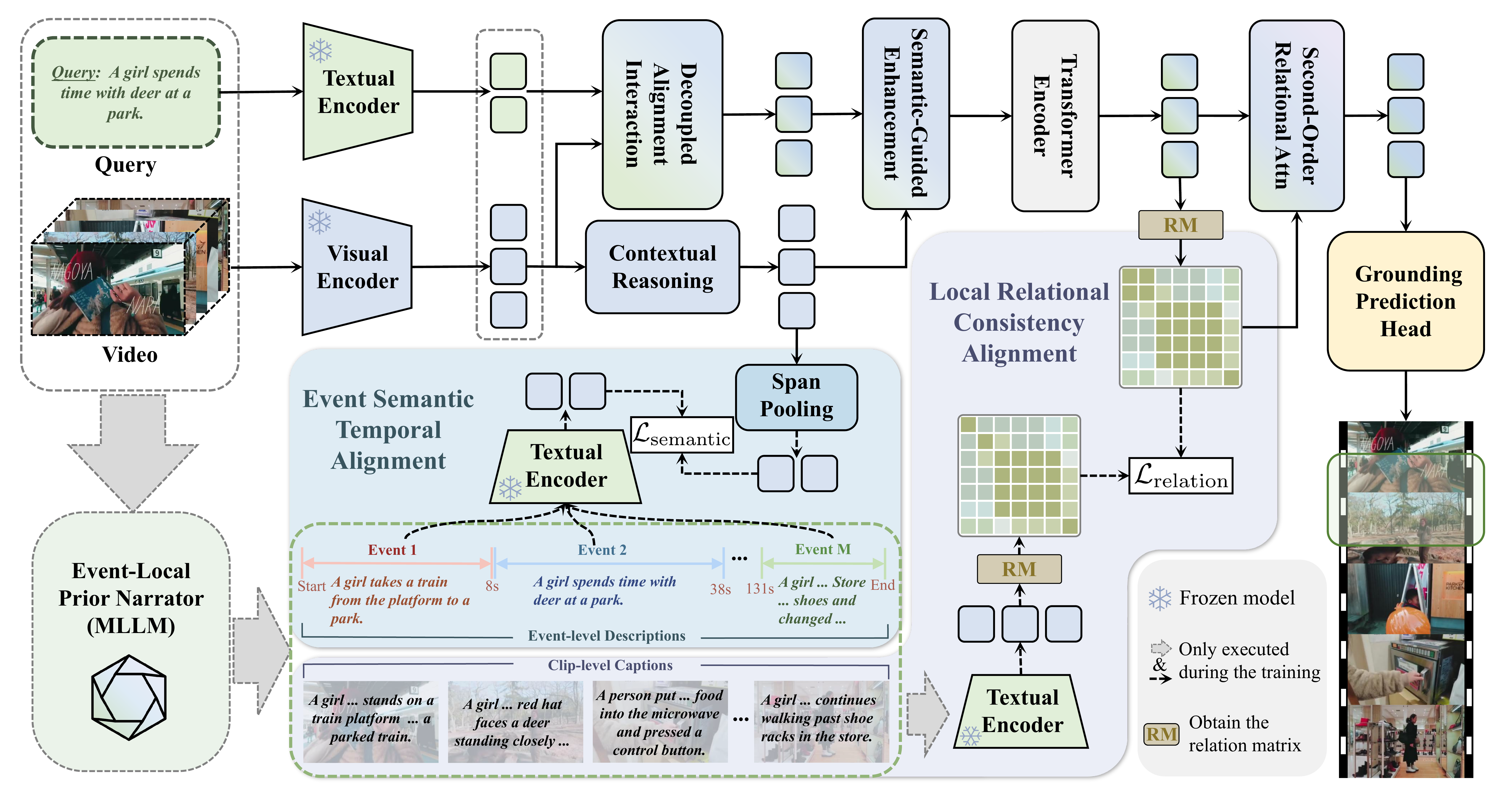}
   \caption{The architecture of the proposed MASRA. Encoders first extract features from a natural-language query and an untrimmed video. Decoupled alignment interaction (DAI) yields interaction features, while a contextual reasoning module produces temporal context, and semantic-guided enhancement fuses them. A Transformer models the fused representation to derive local relations; second-order relational attention further refines clip-to-clip relations; the grounding head predicts start and end timestamps. During training only, an MLLM supplies two textual priors: (1) event spans with descriptions for event semantic temporal alignment, aligning event semantics with pooled temporal context; and (2) clip-level captions to build a relation matrix for local relational consistency alignment, enforcing structural consistency.}
   \label{fig:onecol}
\end{figure*}

\section{Related Work}
\label{sec:formatting}
\subsection{Video Temporal Grounding}
Video temporal grounding aims to localize, in an untrimmed video, the temporal segment that is semantically consistent with a natural language query~\cite{anne2017localizing,gao2017tall, li2022compositional, jung2023overcoming}. Early approaches can be grouped into two technical paradigms. Proposal-based methods generate candidate intervals via sliding windows or temporal anchors and then rank or filter them using cross-modal features~\cite{anne2017localizing,wang2022negative,liu2022skimming}. Proposal-free methods directly regress start-end boundaries or predict frame-level boundary confidence after multimodal interaction~\cite{ning2021interaction, zeng2020dense,lu2019debug,xu2019multilevel}. Recently, Transformer frameworks have been widely adopted for VTG to unify the modeling of global language-video dependencies via query mechanisms~\cite{lei2021detecting,jang2023knowing,sun2024tr,sun2025diversifying}. Subsequent research pursues stronger cross-modal representations~\cite{moon2023correlation,xiao2024bridging}, joint training across tasks~\cite{sun2024tr, jang2023knowing}, and temporal feature mining~\cite{yang2024task, lee2024bam, sun2025diversifying,ran2025kda}. For example, TR-DETR reinforces cross-task feedback under joint temporal grounding \& highlight detection training \cite{sun2024tr}, BAM-DETR improves localization accuracy through boundary alignment~\cite{lee2024bam}, and QD-DETR alleviates misalignment~\cite{moon2023query}. In terms of transfer and adaptation, RGTR introduces region guidance to promote diversified responses across temporal spans and reduce redundancy~\cite{sun2025diversifying}. However, many existing methods rely primarily on direct alignment between the query and the target segment, which tends to draw irrelevant background into the alignment and thereby limits semantic and local discriminability. In contrast, our approach introduces event semantics and local relational supervision, improving alignment reliability and boundary separability.

\subsection{Multimodal Large Language Models}
With the rapid progress of large vision-language models, strong semantic reasoning capability has been widely demonstrated~\cite{li2024llava,liu2024improved,touvron2023llama}. Multimodal large language models (MLLMs) couple a vision model with a large language model to form a unified visual reasoning framework, markedly enhancing cross-modal understanding and generation \cite{dai2023instructblip,chen2024internvl}. At the image and short-event levels, MLLMs provide stable descriptive and comprehension abilities, enabling fine-grained visual QA~\cite{jian2024large}, dense captioning~\cite{li2024improving}, and event detection~\cite{ding2025streammind}. For VTG, recent explorations employ video-oriented MLLMs to parse temporal semantics in a dialog-style manner and then perform grounding or assist training~\cite{qu2024chatvtg}. However, these methods typically require invoking the large model at inference time, incurring heavy computational cost and compromising real-time performance, and their localization granularity can be coarse when analyzing longer videos \cite{guotrace,huang2024vtimellm}. Moreover, some studies involve MLLM for additional supervision to enhance performance~\cite{zhao2024image,wang2025declip}, using MLLM-generated semantic signals to enhance the discriminability and generalization of traditional vision modules, with notable gains in detection and representation learning \cite{fu2025llmdet,liao2025llm,gu2025breaking}. Inspired by these, we leverage MLLM only during training to produce accurate event-level and clip-level text, constructing dual supervision that bridges the cross-modal semantic gap and strengthens both boundary discrimination and temporal consistency, while introducing no inference overhead.

% zhao2024image
% wang2025declip

\section{Methodology}
\subsection{Overall Framework}
Figure~\ref{fig:onecol} presents the architecture of the proposed MASRA. In the grounding backbone, Decoupled Alignment Interaction (DAI) first performs context-aware cross-modal interaction, where a context-aware codebook generates auxiliary tokens to absorb query-irrelevant semantics and stabilize the alignment process, thereby yielding interaction features.  
In parallel, the video features undergo contextual reasoning to produce a temporal context, which is then aligned with event semantics and used to guide the enhancement of interaction features, further yielding temporal features. Based on these features, a pairwise similarity relation is constructed for training-time relational alignment, and the features are further refined accordingly. Finally, a grounding prediction head outputs the target moment.

During training, an MLLM generates clip-level captions and event-level descriptions with temporal spans, based on which two MLLM-assisted alignments are introduced. In Event Semantic Temporal Alignment (ESTA), temporal span pooling converts the temporal context into event features, which are aligned with the corresponding event description features, strengthening the coupling between event semantics and temporal context and improving span discriminability. Based on the semantically aligned context learned through ESTA, semantic-guided enhancement further guides the fusion of interaction features. In Local Relational Consistency Alignment (LRCA), a textual correlation matrix computed from clip-level caption features is regularized against the similarity relation derived from temporal features, enforcing consistency between the relational structure among temporal features and detailed local inter-clip relations. On top of the relation structure learned through LRCA, second-order relational attention further performs lightweight local refinement of the similarity map. Notably, the MLLM and the two MLLM-assisted alignments are used only during training; inference does not rely on the MLLM.

\subsection{Problem Formulation}
Given an untrimmed video \(V = \{c_i\}_{i=1}^{T}\) containing \(T\) sampled clips and a natural language query \(Q = \{w_i\}_{i=1}^{L}\) with \(L\) words, the goal of VTG is to predict the temporal moment \((t_s, t_e)\) that best aligns with the semantic meaning of the query on the time axis, where \(1 \leq t_s < t_e \leq T\). 
Following prior VTG methods, we use frozen video and text encoders to extract initial representations for the video and query. We then employ MLPs to map the features into a shared dimension \(C\). Specifically, the video features are represented as \(\mathcal{V} = \{v_i\}_{i=1}^{T}  \in \mathbb{R}^{T \times C}\), where \(T\) denotes the number of video clips and \(C\) is the feature dimension. Similarly, the text features are represented as \(\mathcal{Q} = \{q_i\}_{i=1}^{L} \in \mathbb{R}^{L \times C}\), where \(L\) is the number of tokens in the query.

\begin{figure}[t]
  \centering
   \includegraphics[width=0.99\linewidth]{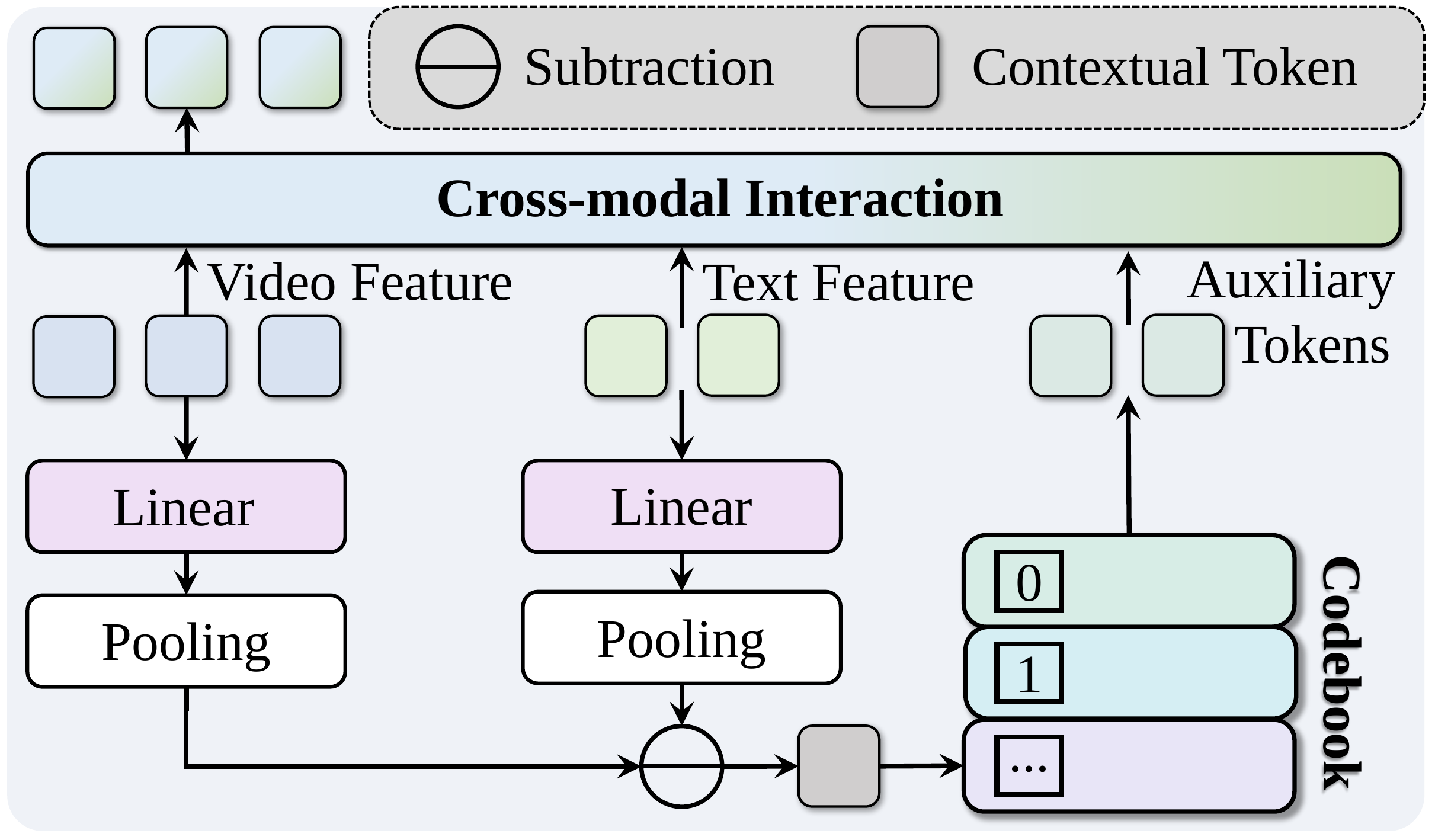}
   \caption{The structure of decoupled alignment interaction. The pooled video and query representations are subtracted to form a contextual token, which is used to retrieve auxiliary tokens from a codebook. The auxiliary tokens then participate in interaction together with video and query features.}
   \label{fig:dai}
\end{figure}

\subsection{Decoupled Alignment Interaction}
We first describe the Decoupled Alignment Interaction module, which serves as the main interaction component of the backbone.
As shown in Figure~\ref{fig:dai}, DAI performs adaptive contextual alignment decoupling and enhances the cross-modal interaction between the video and the query.
First, the video features and the text features are processed through linear transformations and pooling operations to extract their global representations.
Then, subtraction is introduced to compute the differences between the modalities, serving as the background context. This can be formulated as:
\begin{equation}
\mathbf{z} = \text{Pool}(\text{Linear}_v(\mathcal{V})) - \text{Pool}(\text{Linear}_q(\mathcal{Q})).
\label{eq:z}
\end{equation}

Next, we generate \(K\) auxiliary tokens from the codebook conditioned on the context token \(\mathbf{z}\).
Specifically, let the codebook be \(\mathcal{B} = \{\mathbf{b}_k \in \mathbb{R}^{C}\}_{k=1}^{K_B}\), and let
\(\{k_j\}_{j=1}^{K} \subseteq \{1,\dots,K_B\}\) denote the indices of the \(K\) codewords with the smallest distances
\(\|\mathbf{z} - \mathbf{b}_k\|_2\).
The resulting auxiliary token set is \(\mathcal{A} = \{\mathbf{b}_{k_j}\}_{j=1}^{K}\).
The auxiliary tokens, together with the video and query features, participate in the cross-modal interaction to absorb attention from query-irrelevant semantics and stabilize the alignment process, yielding the interaction feature \(\mathcal{I} \in \mathbb{R}^{T \times C}\).

During training, we adopt a VQ-style codebook loss to update the codebook and stabilize the embedding vectors:
\begin{equation}
\mathcal{L}_{\mathrm{cb}}
= \frac{1}{K}\sum_{j=1}^{K} \big\|\mathrm{sg}(\mathbf{z}) - \mathbf{b}_{k_j}\big\|_2^2
+ \beta \big\|\mathbf{z} - \mathrm{sg}(\hat{\mathbf{z}})\big\|_2^2,
\end{equation}
where \(\hat{\mathbf{z}} = \frac{1}{K}\sum_{j=1}^{K} \mathbf{b}_{k_j}\) denotes the aggregated quantized vector, and \(\mathrm{sg}(\cdot)\) denotes the stop-gradient operator.

\subsection{MLLM-Assisted Textual Priors}
During training, MASRA introduces an MLLM-assisted supervision branch.
Given the sequence of sampled clips and the corresponding textual query, the MLLM serves as an event-local prior narrator and produces two types of textual priors: (1) event-level descriptions with temporal spans and (2) clip-level captions.

For event-level priors, the MLLM analyzes the temporal dimension to identify semantically coherent segments.
We parse the output into a structured set
\(\{(y_i,[s_i,e_i])\}_{i=1}^{M}\),
where \(M\) is the number of events, \(y_i\) is the natural language description of the \(i\)-th event, and \([s_i,e_i]\) denote its start and end indices on the sampled clip sequence.
The set contains the annotated ground-truth segment.
Each \(y_i\) is encoded by the textual encoder to obtain event representations
\(\mathcal{O}=\{\mathbf{o}_i\}_{i=1}^{M} \in \mathbb{R}^{M\times D}\),
where \(\mathbf{o}_i\) is the language feature of the \(i\)-th event, and \(D\) is the dimension of the textual encoder.

For clip-level priors, the MLLM generates a detailed caption for each clip (sampled frame), yielding \(T\) texts.
Each caption is fed into the textual encoder to extract the global token or an equivalent descriptive representation, forming the clip-level semantic set
\(\mathcal{C} = \{\mathbf{c}_t\}_{t=1}^{T}\),
where \(\mathbf{c}_t\) denotes the semantic vector of the \(t\)-th clip.

The former provides event-level semantic supervision for ESTA, while the latter is used to derive clip-level relational supervision for LRCA, together providing MLLM-assisted supervision for the grounding backbone.

\subsection{Event Semantic Temporal Alignment}
\label{sec:esta}

Event Semantic Temporal Alignment (ESTA) provides event-level semantic supervision for the temporal context branch of the backbone.
In parallel to Decoupled Alignment Interaction (DAI), the video features are processed by a contextual reasoning module to obtain a temporal context representation
\(\mathcal{H} \in \mathbb{R}^{T \times C}\).
The goal of ESTA is to align this temporal context with event semantics, so that the learned context becomes more semantically localized and more discriminative at the span level.

For the \(i\)-th event span \([s_i, e_i]\), we compute the video-side event representation by a linear layer and mean pooling the temporal context \(\mathcal{H}=\{\mathbf{h}_t\}_{t=1}^{T}\) within the span:
\begin{equation}
\mathbf{u}_i =
\frac{1}{e_i - s_i + 1}
\sum_{t=s_i}^{e_i}\mathrm{Linear}(\mathbf{h}_t),
\end{equation}
where \(e_i - s_i + 1\) denotes the number of clips contained in the corresponding span.
This process yields the model-side event set
\(\mathcal{U}=\{\mathbf{u}_i\}_{i=1}^{M} \in \mathbb{R}^{M\times D}\),
which is in one-to-one correspondence with the text-side event set
\(\mathcal{O}=\{\mathbf{o}_i\}_{i=1}^{M}\).
To explicitly align event semantics with temporal context, we adopt an alignment loss that pulls each \(\mathbf{u}_i\) toward its corresponding \(\mathbf{o}_i\):
\begin{equation}
\mathcal{L}_{\mathrm{semantic}}
=
\frac{1}{M}\sum_{i=1}^{M}\big(1-\cos(\mathbf{u}_i,\mathbf{o}_i)\big),
\end{equation}
where \(\cos(\cdot,\cdot)\) denotes cosine similarity.
This loss binds temporal locality with contextual semantics, enabling the model to discriminate event spans and learn event-aware context.

\begin{figure}[t]
  \centering
   \includegraphics[width=0.99\linewidth]{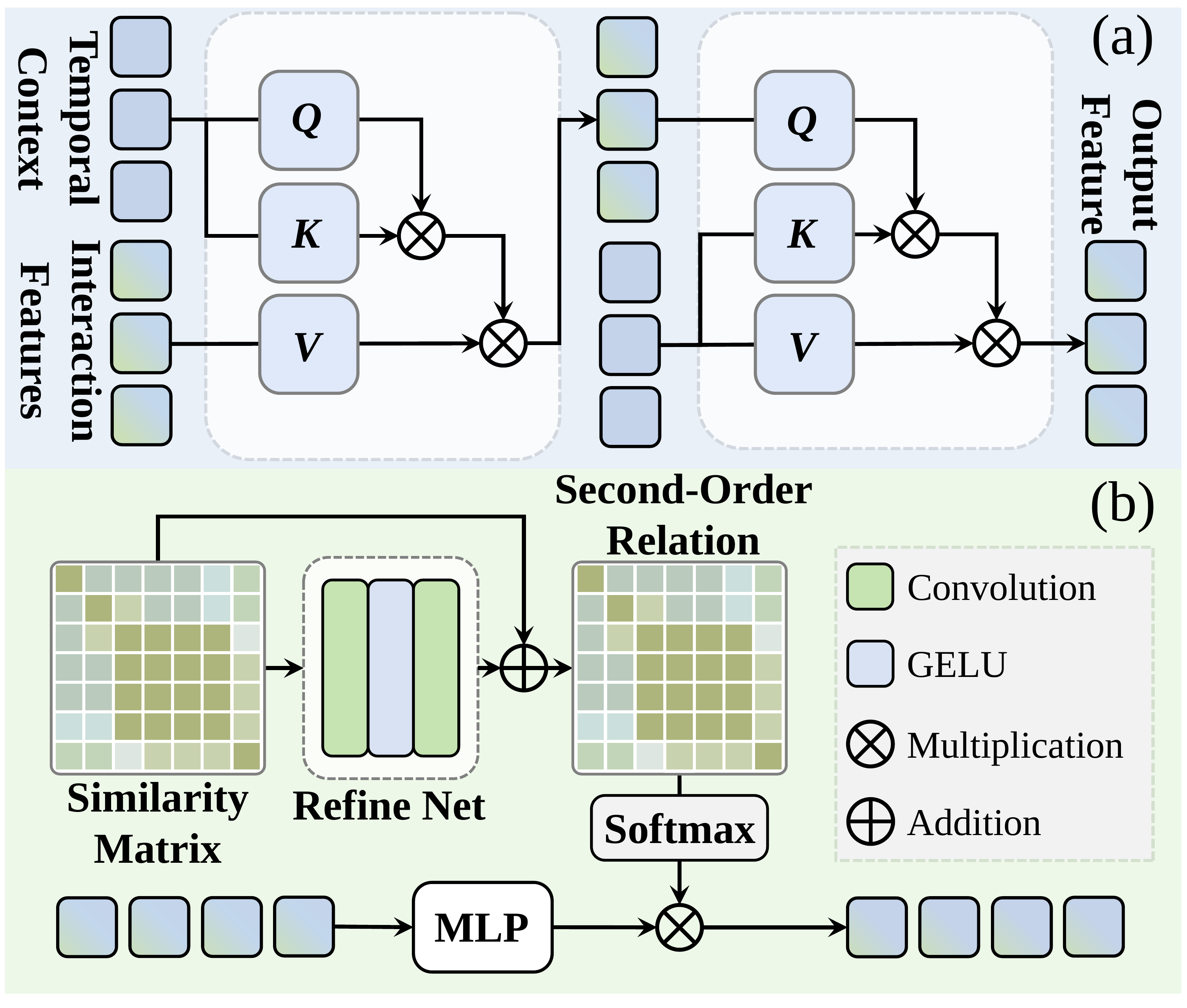}
   \caption{The structure of (a) semantic-guided enhancement and (b) second-order relational attention. SGE uses semantically aligned temporal context to guide feature fusion, while SORA refines the clip-level similarity matrix and feeds the refined relation back to the temporal features.}
   \label{fig:fusion}
\end{figure}

\paragraph{Semantic-Guided Enhancement.}
Based on the semantically aligned temporal context \(\mathcal{H}\), we further employ semantic-guided enhancement to guide feature fusion, as illustrated in Figure~\ref{fig:fusion}(a). Since \(\mathcal{H}\) is directly supervised by ESTA, it provides a cleaner and more semantically reliable temporal structure than the raw interaction feature \(\mathcal{I}\). Accordingly, SGE is not designed as a standard cross-attention block that directly uses one feature source to query another. Instead, it adopts two consecutive attention stages with distinct roles, so that the semantically aligned context first determines where interaction cues should be aggregated, and then further supplements what semantic context should be injected.

Specifically, the first stage uses the temporal structure of \(\mathcal{H}\) to redistribute the interaction feature:
\begin{equation}
\mathcal{E}_1 = \mathrm{Attn}(Q=\mathcal{H}, K=\mathcal{H}, V=\mathcal{I}).
\end{equation}
Here, the aggregation weights are determined by the self-structured temporal affinity of \(\mathcal{H}\), while the aggregated content is drawn from the interaction feature \(\mathcal{I}\), which helps suppress noisy or query-irrelevant interactions and preserve temporally coherent aggregation.
The second stage then injects semantic context back into the redistributed interaction features:
\begin{equation}
\mathcal{E}_2 = \mathrm{Attn}(Q=\mathcal{E}_1, K=\mathcal{H}, V=\mathcal{H}).
\end{equation}
In this way, the first stage performs semantic-guided redistribution of interaction cues, whereas the second stage enriches the redistributed feature with event-aware temporal semantics.
Finally, the fused feature \(\mathcal{E}_2\) is fed into a subsequent Transformer encoder to produce the temporal feature
\(\mathcal{E} \in \mathbb{R}^{T \times C}\).

Overall, ESTA provides event-level semantic supervision for the temporal context branch, and SGE uses the resulting semantically aligned context to guide feature fusion in the backbone.

\begin{table*}[t]
\centering\setlength{\tabcolsep}{8pt}\renewcommand{\arraystretch}{1.0}
\caption{Performance comparison of video temporal grounding on QVHighlights dataset (test and val splits) with SF{+}C features,  \textbf{Bold} and \underline{Underline} indicate the best and second-best results, respectively.}

\begin{tabular}{l  c c c c c  c c c c c}
\toprule 
\multicolumn{1}{l}{\multirow{3}{*}{Method}} & \multicolumn{5}{c}{Test}  & \multicolumn{5}{c}{Val}   \\   
\cmidrule(lr){2-6} \cmidrule(lr){7-11}
& \multicolumn{2}{c}{R1} & \multicolumn{3}{c}{mAP} & \multicolumn{2}{c}{R1} & \multicolumn{3}{c}{mAP} \\
\cmidrule(lr){2-3} \cmidrule(lr){4-6}\cmidrule(lr){7-8} \cmidrule(lr){9-11}
& @0.5 & @0.7 & @0.5  & @0.75 & Avg. & @0.5 & @0.7 & @0.5  & @0.75 & Avg. \\ 
\midrule
M-DETR \cite{lei2021detecting}        & 52.89 & 33.02 & 54.82 & 29.17 & 30.73 & 53.94 & 34.84 &  --   &  --   & 32.20 \\ 
UMT \cite{liu2022umt}                 & 56.23 & 41.18 & 53.83 & 37.01 & 36.12 & 60.26 & 44.26 & 56.70 & 39.90 & 38.59 \\ 
QD-DETR \cite{moon2023query}          & 62.40 & 44.98 & 62.52 & 39.88 & 39.86 & 62.68 & 46.66 & 62.23 & 41.82 & 41.22 \\ 
UniVTG \cite{lin2023univtg}           & 58.86 & 40.86 & 57.60 & 35.59 & 35.47 & 59.74 &   --  &  --   &  --   & 36.13 \\ 
MomentDiff \cite{li2024momentdiff}    & 57.42 & 39.66 & 54.02 & 35.73 & 35.95 &   --  &  --  &  --   &  --   &  --   \\ 
CG-DETR \cite{moon2023correlation}    & 65.43 & 48.38 & 64.51 & 42.77 & 42.86 & 67.35 & 52.06 & 65.57 & 45.73 & 44.93 \\
UVCOM \cite{xiao2024bridging}         & 63.55 & 47.47 & 63.37 & 42.67 & 43.18 & 65.10 & {51.81} &  --   &  --   & 45.79 \\ 
BAM-DETR \cite{lee2024bam}            & 62.71 & 48.64 & 64.57 & \underline{46.33} & 45.36 & 65.10 & 51.61 & 65.41 & \underline{48.56} & \underline{47.61} \\ 
TR-DETR \cite{sun2024tr}              & 64.66 & 48.96 & 63.98 & 43.73 & 42.62 & {67.10} & 51.48 & {66.27} & 46.42 & 45.09 \\ 
TaskWeave \cite{yang2024task} &   --  &  --  &  --   &  --   &  --  & 64.26 & 50.06 & 65.39 & 46.47 & 45.38 \\
KDA \cite{ran2025kda}                 & \textbf{66.70} & \underline{50.88} & \underline{67.57} & 46.31 & \underline{45.67} & \textbf{69.11} & \underline{53.46} & \underline{68.17} & 48.04 & 47.41 \\ 
% R2-Tuning \cite{liu2024r2tuning}      & \underline{68.03} & 49.35 & \underline{69.04} & 47.56 & 46.17 & 68.71 & 52.06 &  --   &  --   & 47.59 \\
RGTR \cite{sun2025diversifying}               & {65.50} & 49.22 & 67.12 & 45.77 & 45.53 & 67.68 & 52.90 & 67.38 & 48.00 & 46.95 \\

MASRA (Ours)                       & \underline{66.23} & \textbf{50.90} & \textbf{68.16} & \textbf{48.58} & \textbf{47.94} &  
\underline{68.97} & \textbf{54.19} & \textbf{68.93} & \textbf{52.47} & \textbf{50.97} \\
\bottomrule 
\end{tabular} 
\label{tab:qvhl}
\end{table*}

% \underline{68.92} & \textbf{55.17} & \textbf{69.39} & \textbf{53.85} & \textbf{52.09} \\

\subsection{Local Relational Consistency Alignment}
\label{sec:lrca}

On top of the fused temporal feature \(\mathcal{E}\), LRCA further constrains the similarity relation at the clip level.
Rather than only learning which span matches the query, LRCA regularizes how clips should relate to each other locally, so as to improve local temporal consistency and boundary separability.

We first construct the clip-level visual similarity matrix $S$ from the temporal feature \(\mathcal{E}\):
\begin{equation}
s_{ij} =
\frac{\mathbf{e}_i^{\top}\mathbf{e}_j}{\|\mathbf{e}_i\|_2\,\|\mathbf{e}_j\|_2},
\qquad
S=[s_{ij}]_{i,j=1}^{T}\in\mathbb{R}^{T\times T},
\end{equation}
where \(\mathbf{e}_i\) denotes the \(i\)-th feature in \(\mathcal{E}\).

On the textual side, each clip-level caption is encoded into a semantic feature \(\mathbf{c}_t\), and the textual relation matrix \(R=[r_{ij}]_{i,j=1}^{T}\in\mathbb{R}^{T\times T}\) is constructed by computing the pairwise cosine similarity between all clip-level caption features. In this way, each entry \(r_{ij}\) measures the semantic relatedness between the captions of the \(i\)-th and \(j\)-th clips, providing a text-derived local relational structure over the video sequence.
We then impose a consistency constraint that fits the visual relation to the textual relation structure:
\begin{equation}
\mathcal{L}_{\mathrm{relation}}
=
\frac{1}{T^{2}}
\sum_{i=1}^{T}\sum_{j=1}^{T}
\big\|s_{ij}-r_{ij}\big\|_{2}^{2},
\end{equation}
where \(s_{ij}\) and \(r_{ij}\) are the \((i,j)\)-th entries of \(S\) and \(R\), respectively.
This constraint operates directly on the relational structure rather than on individual clip embeddings.
It enables the model to judge which clips should cluster together and which should be separated in the similarity structure, thereby improving local discriminability and boundary discriminability.

\paragraph{Second-Order Relational Attention.}
After LRCA has regularized the clip-level similarity structure, we further employ second-order relational attention to refine the learned similarity map, as illustrated in Figure~\ref{fig:fusion}(b).
We treat SORA as a lightweight and effective local refinement module rather than a long-range graph propagation mechanism in practice.

Specifically, SORA refines the similarity matrix with a lightweight refine network to obtain a second-order similarity matrix:
\begin{equation}
\widetilde{S} = S + \phi(S),
\end{equation}
where \(\phi(\cdot)\) denotes the refine network composed of convolutions and activations.
This refinement captures neighboring second-order relational patterns on the similarity map and suppresses local noise.
We then normalize \(\widetilde{S}\) and reweight the temporal features to produce the refined feature
\(\mathcal{F}\in\mathbb{R}^{T\times C}\):
\begin{equation}
\mathcal{F} = \mathrm{softmax}(\widetilde{S}) \cdot \mathrm{MLP}(\mathcal{E}).
\end{equation}
This operation sharpens temporal structure and improves local boundary sensitivity.

In this way, LRCA regularizes the relational structure using clip-level textual priors, while SORA further refines that structure and feeds it back into the temporal features.

\subsection{Moment Prediction and Training}
We adopt a prediction scheme consistent with prior VTG work~\cite{cao2025flashvtg, pujol2025sparse, zhao2025ld}.
From the refined temporal features \(\mathcal{F}\), we form grounding tokens and construct a lightweight prediction head to regress the start and end timestamps together with the corresponding classification confidence scores, while a separate saliency branch outputs clip-level saliency predictions.

During training, we combine multiple constraints:
\begin{equation}
\begin{aligned}
\mathcal{L}_{\mathrm{overall}}
&= \mathcal{L}_{\mathrm{vtg}}
+ \lambda_{\mathrm{sal}}\,\mathcal{L}_{\mathrm{sal}}
+ \lambda_{\mathrm{sem}}\,\mathcal{L}_{\mathrm{semantic}} \\
&\quad + \lambda_{\mathrm{rel}}\,\mathcal{L}_{\mathrm{relation}}
+ \lambda_{\mathrm{cb}}\,\mathcal{L}_{\mathrm{cb}},
\end{aligned}
\end{equation}
where \(\mathcal{L}_{\mathrm{vtg}}\) denotes the moment retrieval loss using a classification \(+\) \(L_1\) \(+\) GIoU combination, \(\mathcal{L}_{\mathrm{sal}}\) denotes the clip-level saliency loss, \(\mathcal{L}_{\mathrm{semantic}}\) and \(\mathcal{L}_{\mathrm{relation}}\) are the two MLLM-assisted alignment losses, and \(\mathcal{L}_{\mathrm{cb}}\) is the codebook loss.
The coefficients \(\lambda_{\mathrm{sal}}\), \(\lambda_{\mathrm{sem}}\), \(\lambda_{\mathrm{rel}}\), and \(\lambda_{\mathrm{cb}}\) are balancing parameters.

\begin{table*}[t]
\centering\setlength{\tabcolsep}{10pt}\renewcommand{\arraystretch}{1.0}
\caption{Performance comparison of video temporal grounding on TACoS and Charades-STA datasets with SF{+}C features.  \textbf{Bold} and \underline{Underline} indicate the best and second-best results, respectively.}
\begin{tabular}{l cccc cccc}
\toprule
\multirow{3}{*}{Method} & \multicolumn{4}{c}{TACoS} & \multicolumn{4}{c}{Charades-STA} \\
\cmidrule(lr){2-5}\cmidrule(lr){6-9}
& R1@0.3 & R1@0.5 & R1@0.7 & mIoU & R1@0.3 & R1@0.5 & R1@0.7 & mIoU \\
\midrule
2D-TAN \cite{zhang2020learning}       & 40.01 & 27.99 & 12.92 & 27.22 & 58.76 & 46.02 & 27.50 & 41.25 \\
VSLNet \cite{zhang2020span}           & 35.54 & 23.54 & 13.15 & 24.99 & 60.30 & 42.69 & 24.14 & 41.58 \\
M-DETR \cite{lei2021detecting}        & 37.97 & 24.67 & 11.97 & 25.49 & 65.83 & 52.07 & 30.59 & 45.54 \\
UniVTG \cite{lin2023univtg}           & 51.44 & 34.97 & 17.35 & 33.60 & 70.81 & 58.01 & 35.65 & 50.10 \\
UVCOM \cite{xiao2024bridging}         &   --  & 36.39 & 23.32 &  --   &  --   & 59.25 & 36.64 &  --   \\
% R2-Tuning \cite{liu2024r2tuning}      & 49.71 & 38.72 & 25.12 & 35.92 & 70.91 & 59.78 & 37.02 & 50.86 \\

CG-DETR \cite{moon2023correlation}    & 52.23 & 39.61 & 22.23 & 36.48 & 70.43 & 58.44 & 36.34 & 50.13 \\

KDA \cite{ran2025kda}                           &   --  & 40.13 & \underline{24.34} &  --   &  --   & \underline{60.23} & \underline{37.63} &  --   \\
RGTR \cite{sun2025diversifying}               & \underline{53.04} & \underline{40.31} & 24.32 & \underline{37.44} & \textbf{72.04} & 57.93 & 35.16 & \underline{50.32} \\

MASRA (Ours)                           & \textbf{54.61} & \textbf{41.97} & \textbf{26.17} & \textbf{38.84} & \underline{72.01} & \textbf{61.08} & \textbf{38.75} & \textbf{51.22} \\
\bottomrule
\end{tabular}
\label{tab:tacos_charades_sfc}
\end{table*}

% =============================== VGG-only（两指标 + Feat.） ===============================
\begin{table}[t]
\centering\setlength{\tabcolsep}{8pt}\renewcommand{\arraystretch}{1.0}
\caption{Performance comparison of video temporal grounding on {Charades-STA} dataset with the VGG feature.}
\begin{tabular}{l|c cc}
\toprule
Method & Feat. & R1@0.5 & R1@0.7 \\
\midrule
2D-TAN \cite{zhang2020learning}       & VGG & 40.94 & 22.85 \\
FVMR \cite{gao2021fast}               & VGG & 42.36 & 24.14 \\
UMT \cite{liu2022umt}       & VGG & 48.31 & 29.25 \\
MomentDiff \cite{li2024momentdiff}    & VGG & 51.94 & 28.25 \\
QD-DETR \cite{moon2023query}          & VGG & 52.77 & 31.13 \\
TR-DETR \cite{sun2024tr}              & VGG & 53.47 & 30.81 \\
TaskWeave \cite{yang2024task}               & VGG & \underline{56.51} & {33.66} \\
KDA \cite{ran2025kda}                 & VGG & {55.36} & \underline{34.50} \\
RGTR \cite{sun2025diversifying}                 & VGG & {55.48} & {34.33} \\

MASRA (Ours)                           & VGG & \textbf{57.52} & \textbf{38.46} \\
\bottomrule
\end{tabular}
\label{tab:vgg_charades}
\end{table}

% 57.03 39.84

\section{Experiments}
\subsection{Datasets and Metrics}
\noindent\textbf{Datasets.}
We evaluate on three widely used temporal grounding benchmarks: {QVHighlights}~\cite{lei2021detecting}, {Charades-STA}~\cite{gao2017tall}, and {TACoS}~\cite{regneri2013grounding}. 
QVHighlights pairs YouTube videos with natural-language queries and provides multiple ground-truth moments per query, enabling fine-grained evaluation.
Charades-STA extends the Charades dataset with 16{,}128 sentence-moment annotations over 9{,}848 indoor videos, and we follow the standard split with 12{,}408 training and 3{,}720 testing pairs~\cite{moon2023query}.
TACoS consists of long cooking videos with dense temporal language annotations, posing challenges for precise localization under long-form temporal context.
We adopt the standard public splits and evaluation used in prior works~\cite{moon2023query,sun2024tr}.

\noindent\textbf{Metrics.}
Following standard practice in VTG~\cite{lei2021detecting,sun2024tr,xiao2024bridging, pujol2025sparse}, we report grounding metrics.
For {QVHighlights}, we evaluate Recall@1 (R1) at IoU thresholds $0.5$ and $0.7$, mean Average Precision (mAP) at IoU $0.5$ and $0.75$, and the averaged mAP over IoU thresholds $[0.5\!:\!0.05\!:\!0.95]$ for comprehensive comparison.
For {Charades-STA} and {TACoS}, we report R1 at IoU thresholds $0.3$, $0.5$, and $0.7$, together with the mean IoU of the top-1 prediction.

\subsection{Implementation Details}
Following prior practice, we adopt SlowFast~\cite{feichtenhofer2019slowfast} and CLIP~\cite{radford2021learning} (SF+C) as the video encoders, and encode text queries with CLIP's text encoder.
For the Charades\mbox{-}STA dataset, we additionally report variants that use VGG~\cite{simonyan2014very} as the video encoder and GloVe~\cite{pennington2014glove}  for text embeddings.
For the MLLM that generates auxiliary captions, we employ GPT\mbox{-}5~\cite{openai2025gpt5} and obtain features by encoding the generated texts with the textual encoder of the model. We preprocess the textual features of the captions and use them directly during training.
In the decoupled alignment interaction module, the codebook size is set to \(K_{B}=1024\), and we retrieve the top\mbox{-}50 auxiliary tokens at each selection step.
We optimize the network using AdamW~\cite{loshchilov2019decoupled} with a learning rate of \(1\times 10^{-4}\), weight decay of \(1\times 10^{-4}\), a batch size of 32, and train for 400 epochs. We do not use any additional pretraining on other data beyond the adopted frozen encoders
All experiments are conducted on one NVIDIA A100 GPU.

% ===================== Ablation：对齐分支（ECTA / LRCA） =====================
\begin{table}[t]
\centering
\caption{Ablation study of the two MLLM-assisted alignments: event semantic temporal alignment (ESTA) and local relational consistency alignment (LRCA).}
\setlength{\tabcolsep}{8pt}
\begin{tabular}{cc|ccc}
\toprule
\multirow{1}{*}{ESTA} &  \multirow{1}{*}{LRCA}  & R1@0.5 & R1@0.7 & mAP \\ 
\midrule
       &           & 65.70 & 52.99 & 48.62 \\ 
\ding{51} &        & 67.64 & 53.76 & 49.84 \\ 
       & \ding{51} & 68.08 & 53.51 & 49.67 \\ 
\ding{51} & \ding{51} & \textbf{68.97} & \textbf{54.19} & \textbf{50.97} \\
\bottomrule
\end{tabular}
\label{tab:ablation_align}
\end{table}

\subsection{Performance Comparison}
Table~\ref{tab:qvhl} reports results on QVHighlights for both the test and val splits. Under the SlowFast+CLIP feature setting, MASRA achieves the best performance on most metrics, except for R1@0.5 on both the test and val splits, where it ranks second. Notably, the mean mAP improves on both splits. These gains validate that our training-time MLLM-assisted alignments effectively bridge the semantic temporal gap and enhance boundary discrimination, leading to substantial improvements in grounding performance.
Table~\ref{tab:tacos_charades_sfc} summarizes the results on TACoS and Charades-STA.
On the long-form TACoS benchmark, our method achieves clear advantages at all metrics, indicating stronger temporal modeling and better boundary stability. 
On Charades-STA, the proposed MASRA achieves the best results on most metrics and ranks second only on R1@0.3.
To verify robustness under legacy backbones, we evaluate our model on Charades-STA using VGG features, as shown in Table~\ref{tab:vgg_charades}. MASRA still attains the best R1@0.5 and R1@0.7, showing that our alignment supervision and backbone designs are complementary and not tied to a specific encoder.

% ===================== Ablation：主干模块（DAI / CGE / SORA） =====================
\begin{table}[t]
\centering
\caption{Ablation study on backbone components of the proposed MASRA, including Decoupled Alignment Interaction (DAI), Semantic-Guided Enhancement (SGE), and Second-Order Relational Attention (SORA).}
\setlength{\tabcolsep}{9pt}\renewcommand{\arraystretch}{1.0}
\begin{tabular}{l|ccc}
\toprule
Setting  & R1@0.5 & R1@0.7 & mAP \\ 
\midrule
% w/o DAI            & 67.58 & 53.94 & 50.73 \\
w/o DAI            & 68.79 & 53.96 & 50.40 \\
% w/o SGE          & 68.46 & 54.74 & 51.67 \\
w/o SGE            & 68.56 & 53.79 & 50.44 \\
% w/o SORA         & 68.24 & 54.33 & 50.92 \\
w/o SORA           & 68.45 & 53.67 & 49.84 \\
Full Model         & \textbf{68.97} & \textbf{54.19} & \textbf{50.97} \\
\bottomrule
\end{tabular}
\label{tab:ablation_backbone}
\end{table}

\subsection{Ablation Studies}
We conduct several ablation studies on the QVHighlights val split under the SF{+}C feature to validate MASRA.

\noindent\textbf{Effect of MLLM-assisted alignment.}
Table~\ref{tab:ablation_align} presents the isolated contributions of ESTA and LRCA. 
Each branch brings a clear and consistent performance gain when used alone, improving both recall and mAP over the baseline, and the full configuration yields the largest overall improvement, indicating that event-level semantic temporal alignment and clip-level relational consistency are complementary rather than merely redundant. 
This further confirms that injecting MLLM-derived event semantics and local relational priors during training effectively enhances the discriminability of the learned temporal representations.

\noindent\textbf{Backbone components.}
Table~\ref{tab:ablation_backbone} evaluates the proposed DAI, SGE, and SORA. Removing DAI and directly interacting with video and text features degrades performance, indicating that the context-aware codebook is necessary for absorbing query-irrelevant semantics during cross-modal interaction. Removing SGE, where fusion is replaced by concatenation followed by a linear layer, shows that SGE further improves results by integrating the interaction features with the ESTA-supervised context. Finally, SORA provides additional gains by refining second-order relations and feeding them back into the features.

\begin{figure}[t]
  \centering
   \includegraphics[width=0.99\linewidth]{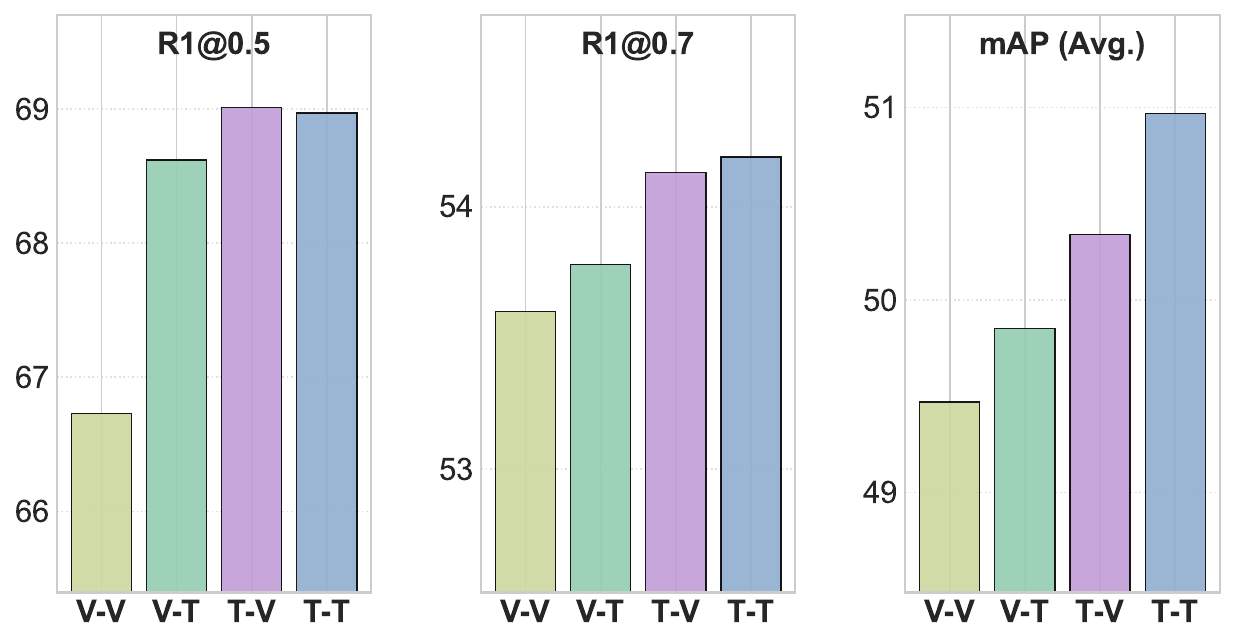}
   \caption{Ablation study on the modality source of alignment priors, where each pair is ordered as ESTA-LRCA and \textit{V}/\textit{T} denote visual/textual priors.}
   \label{fig:ablation1}
\end{figure}

\noindent\textbf{Source of alignment priors.}
Figure~\ref{fig:ablation1} examines the source of alignment priors for ESTA and LRCA. For each branch, we replace the original text-based prior with either visual features (V) or MLLM-generated captions (T), yielding four variants (ESTA-LRCA): V-V, V-T, T-V, and T-T. 
The fully visual setting V-V performs the worst, while introducing text on only one branch (V-T or T-V) brings improvements, and the fully text-based setting T-T achieves the best overall results, showing that MLLM-generated language provides more effective priors and better reduces the cross-modal gap.

% 实际结果，一个方法竖着！：
% methods = ['V-V','V-T','T-V' ,'T-T']

% metrics = {
%     'R1@0.5':     [66.73, 68.62, 69.01, 68.97],
%     'R1@0.7':     [53.60, 53.78, 54.13, 54.19],
%     'mAP (Avg.)': [49.47, 49.85, 50.34, 50.97],
% }

% \caption{Ablation study on the different modal source of alignment priors, where \textit{V} and \textit{T} denote visual features and MLLM-generated textual priors, respectively.}

\begin{figure}[t]
  \centering
   \includegraphics[width=0.99\linewidth]{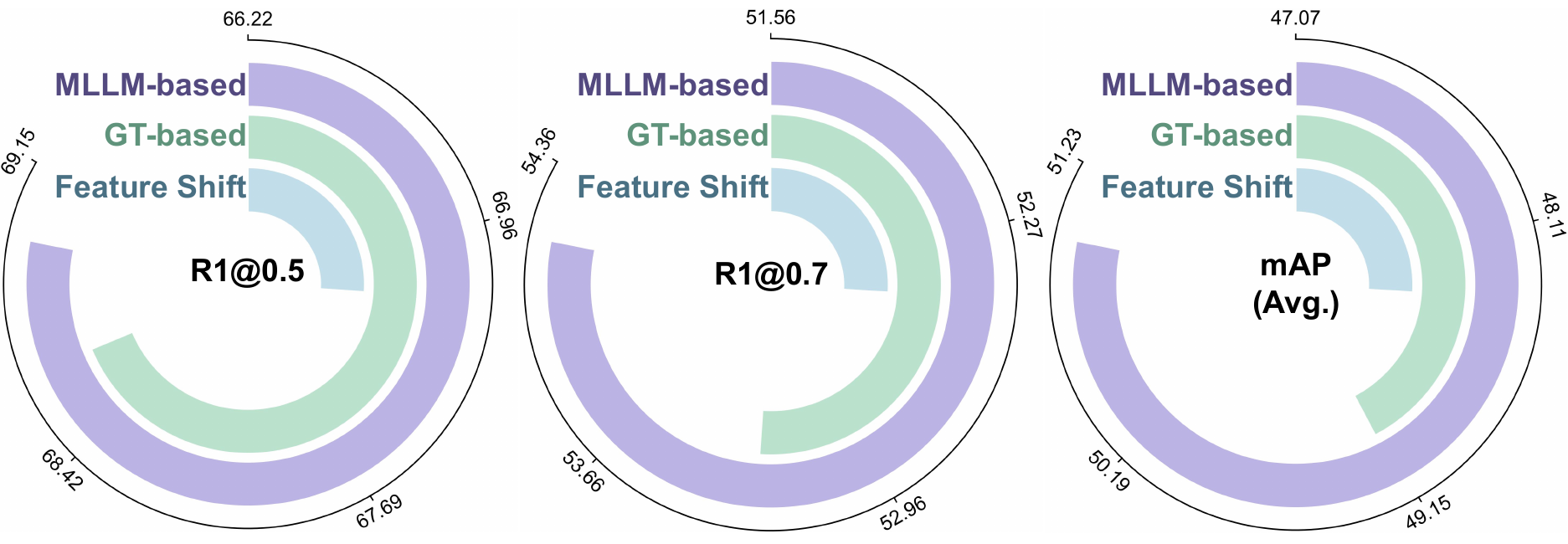}
   \caption{Ablation study of the event span strategies for ESTA, including MLLM-based event analysis, only GT-based spans, and partitions based on feature changes (Feature Shift).}
   \label{fig:ablation2}
\end{figure}
% models = [
%     "MLLM-based ",
%     "GT-based",
%     "Feature Shift",
% ]

% # 每个指标一个列表（顺序与 models 一一对应）
% metrics = {
%     "R1@0.5":       [68.97, 68.64, 67.14],
%     "R1@0.7":       [54.19, 53.08, 52.44],
%     "mAP (Avg.)":   [50.97, 49.18, 48.37],
% }

\noindent\textbf{Event-span sourcing for ESTA.}
Figure~\ref{fig:ablation2} investigates strategies for extracting event spans: an MLLM-based analysis, a GT-only variant that directly uses ground-truth spans, and a feature shift segmentation that partitions sequences by changes in features. The feature shift performs the worst due to imprecise segmentation. Although GT provides correct spans, it lacks dense supervisory signals and thus brings limited gains. In contrast, the MLLM-based approach achieves the best results, as its local event analysis supplies stronger supervision for ESTA.

\subsection{Qualitative Analysis}
\label{sec:qual}
\begin{figure}[t]
  \centering
   \includegraphics[width=0.99\linewidth]{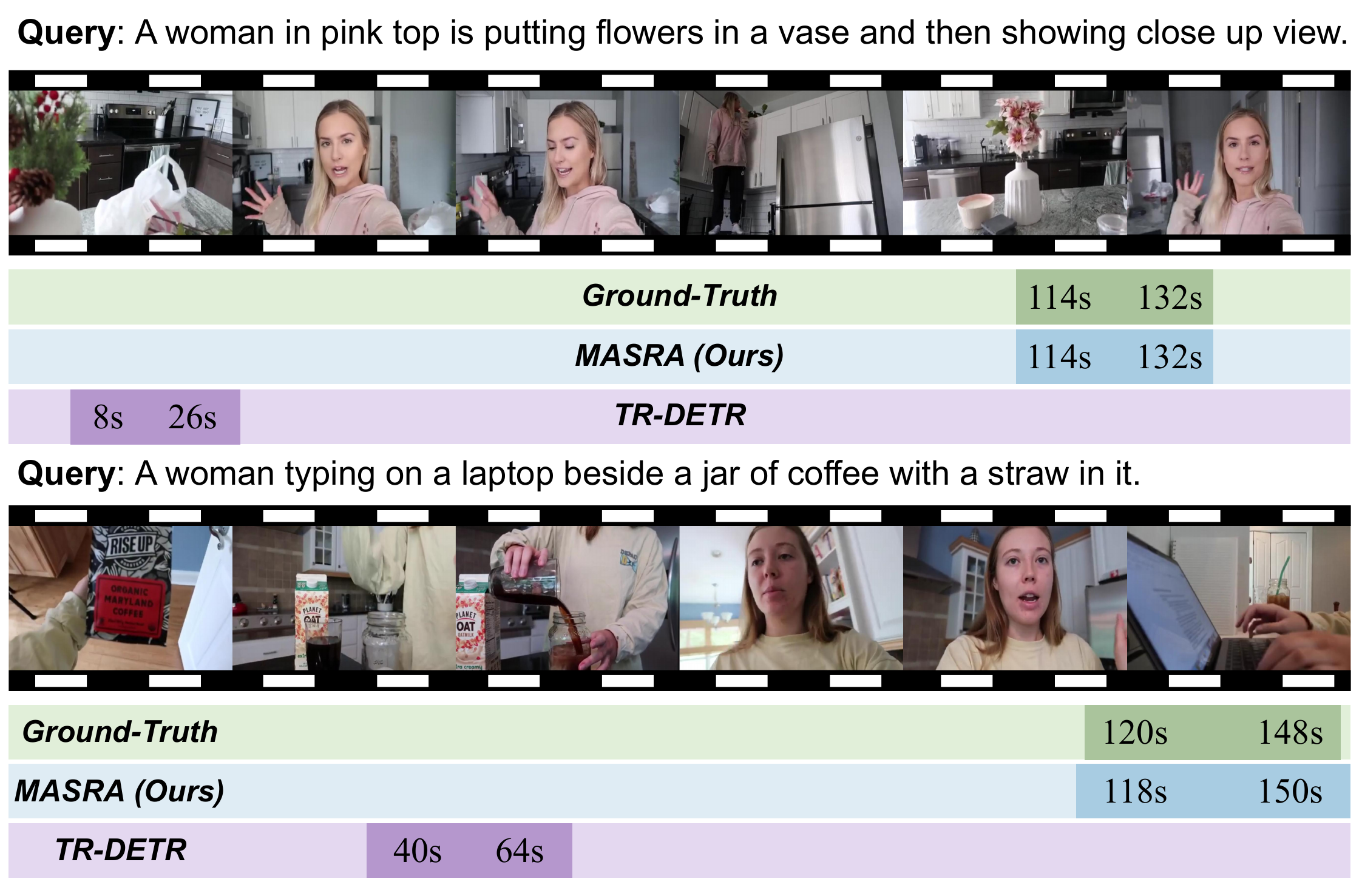}
   \caption{Qualitative comparison on representative examples from the val split of QVHighlights dataset for video temporal grounding task.}
   \label{fig:exp1}
\end{figure}

As shown in Figure~\ref{fig:exp1}, the proposed MASRA more precisely retrieves the temporal moments most relevant to the query, producing grounding predictions that are closer to the ground truth than TR-DETR. This indicates that our method effectively narrows the semantic gap and strengthens boundary discrimination.

\begin{figure}[t]
  \centering
   \includegraphics[width=0.99\linewidth]{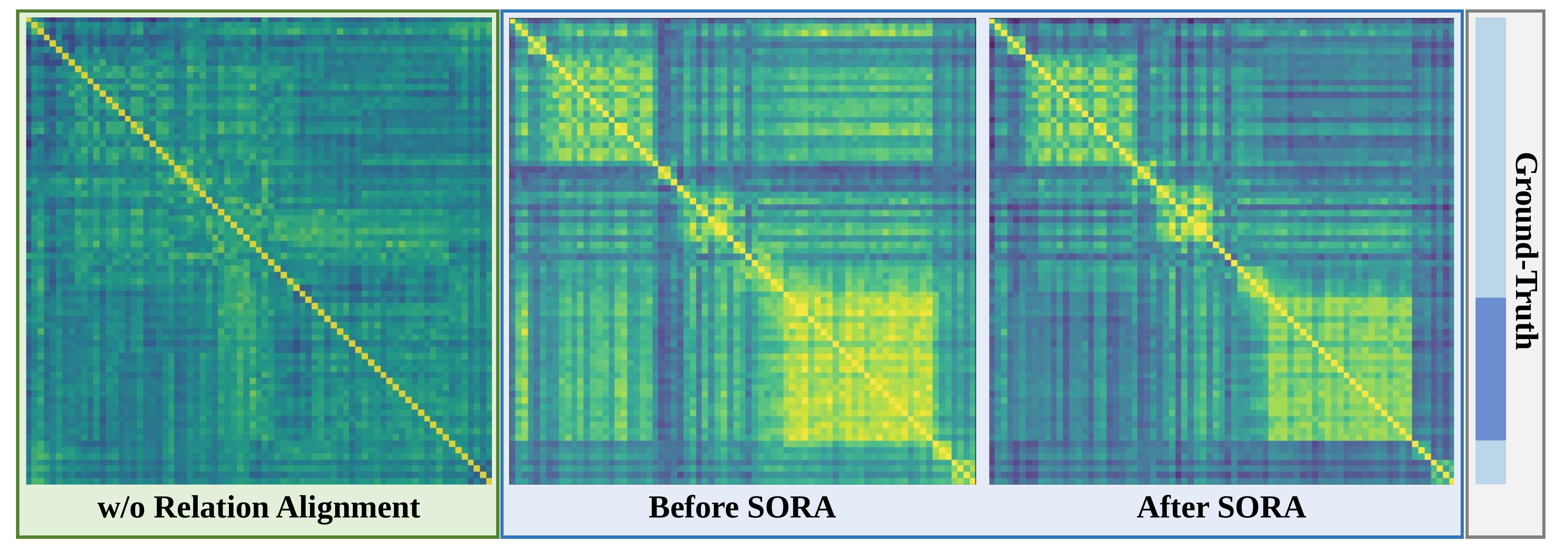}
    \caption{Visualization of clip-level similarity matrices under different relation modeling settings and stages. Left: without local relational consistency alignment. Middle: with LRCA before SORA refinement. Right: with LRCA after SORA refinement. Far-right bar: temporal ground truth.}
   \label{fig:exp2}
\end{figure}

Figure~\ref{fig:exp2} visualizes the similarity matrix: w/o LRCA (left), w/ LRCA (middle \& right). Without LRCA, the matrix exhibits no clear structural pattern. After introducing LRCA, the model already learns coherent local temporal associations and initial block-like boundary structures before SORA. Building on this, SORA performs second-order propagation and reweighting on the similarity matrix, further refining relations and suppressing noise, which yields clearer clusters and boundaries.

\section{Conclusion}
In this paper, we presented MASRA, an MLLM-assisted semantic-relational consistent alignment framework for VTG, which leverages MLLM-generated event-level and clip-level textual priors during training to provide denser supervision for temporal grounding. Based on these priors, we introduced two complementary alignments: event semantic temporal alignment, which strengthens semantic-temporal correspondence at the event level, and local relational consistency alignment, which regularizes clip-level relational structure to improve temporal consistency and boundary separability. We further incorporated decoupled alignment interaction for more adaptive cross-modal alignment, together with two supporting modules that help the backbone better utilize the learned semantic context and relational structure. Since the MLLM is used only during training, the proposed framework introduces no additional inference-time overhead. Extensive experiments on multiple benchmarks demonstrate the overall effectiveness of MASRA and show that it consistently outperforms existing methods.

%%
%% The next two lines define the bibliography style to be used, and
%% the bibliography file.
\bibliographystyle{ACM-Reference-Format}
\bibliography{sample-base}

%%
%% If your work has an appendix, this is the place to put it.
% \appendix

\end{document}